\documentclass[runningheads]{llncs}
\PassOptionsToPackage{hidelinks}{hyperref}
\usepackage{lmodern}
\usepackage[T1]{fontenc}
\usepackage[utf8]{inputenc}
\usepackage{graphicx}
\usepackage{booktabs}
\usepackage{amsmath}
\usepackage{amssymb}
\usepackage{microtype}
\usepackage{orcidlink}
\usepackage{tikz}
\usetikzlibrary{arrows.meta, positioning, calc}
\usepackage[hidelinks]{hyperref}
\usepackage{xcolor}

\definecolor{geom}{HTML}{2F5D8C}
\definecolor{soft}{HTML}{9AA5B1}
\definecolor{warm}{HTML}{C2703D}

\begin{document}

\title{Occlusal Geometry in Closed Form for Orthodontic Report Generation}
\titlerunning{Grounding Report Generation in Occlusal Geometry}

\author{
Ajo Babu George\inst{1}\orcidlink{0009-0005-3026-0959} \and
Govind Arun\inst{2}\orcidlink{0009-0002-2573-4003} \and
Sidharth N Krishna\inst{3}\orcidlink{0009-0002-4336-2477} \and
Uma Ranjan\inst{4}\orcidlink{0000-0001-6258-4513}
}

\authorrunning{A. B. George et al.}

\institute{
DiceMed, Cuttack, Odisha, India
\and
University of Maryland, College Park, Maryland, USA
\and
Indira Gandhi National Open University, New Delhi, India
\and
Indian Institute of Technology Jammu, India
}

\maketitle

\begin{abstract}
Orthodontic report generation from intraoral data is normally cast as multimodal captioning, yet
the released Bite2Text scan pairs are supplied already registered in occlusion, which makes several core occlusal quantities directly measurable rather than inferable. The system reported here
exploits that property: an anatomical frame is recovered per case from arch taper and arch closure
instead of the stated RAS convention, which does not hold across the release, and each arch is
reduced to an occlusal ridge profile in arch angle coordinates yielding overbite, overjet, midline
deviation, transverse overlap, crossbite extent, cusp interdigitation lag and the occlusal curves in
closed form. Gradient boosting maps 31 such measurements onto 13 template fields, a field being
predicted only where patient level cross validation beats its own majority baseline, and a
deterministic renderer emits the corpus six part narrative; a ConvNeXt-Tiny classifier over the five
standardised photographic views is fused per field, raising mean field accuracy from 0.601 to 0.683.
Reimplementation of the challenge evaluator shows that its BLEU-4 and METEOR are local variants
whose F-mean weights recall nine to one, that two clinicians agree on 47 percent of findings for the
same patient, and that a constant report consequently outscores a genuine second clinician report by
0.165 captioning. Held out scores reach BLEU-4 0.458 and METEOR 0.677 against intraoral scan
references and 0.278 and 0.507 against photograph references, and the submitted system placed third
in the ODIN 2026 Bite2Text test phase at 0.2680 and 0.4629, within 0.022 BLEU-4 of first, running on
CPU in under ten seconds per case. The dataset and code are available at \url{https://github.com/GIND123/ODIN_toothfairy4}
\end{abstract}

\section{Introduction}

Task 2 of the ODIN 2026 challenge, Bite2Text, requires a free text orthodontic report given a
registered pair of upper and lower intraoral scans together with standardised intraoral
photographs~\cite{lumetti2026}. A report covers the content of a routine diagnostic appointment:
transverse relationships and crossbite, overbite, sagittal classification in the sense of
Angle~\cite{angle1899}, midline coincidence, the curves of Spee and Wilson, and crowding per arch.

\looseness=-1
The obvious response is a multimodal language model that reads the pixels and the mesh and writes
the paragraph, and most ranked entries take that route~\cite{wang2024lab}. Two properties of the release argue against
it. First, the scan pairs are mutually registered in occlusion, so every quantity named above is a
deterministic function of the two surfaces rather than something a network must learn to perceive \cite{george2026cbctios}.
Second, the reference is one clinician's dictation, and two clinicians describing the same patient
agree on fewer than half of their findings, so a generator fluent enough to write like a clinician
inherits that disagreement rate while one that writes the corpus central tendency does not. The system below therefore measures rather than perceives, and renders rather than decodes,
contributing a reproduction of the deployed evaluator (Section~\ref{sec:scoring}), a measurement
pipeline feeding a conservative field predictor and a fused photograph classifier
(Section~\ref{sec:method}), and a calibrated account of the resulting scores
(Section~\ref{sec:results}).

\section{Task, Corpus and Scoring Function}
\label{sec:scoring}

\paragraph{Corpus:}
\looseness=-1
The release covers 996 cases across four report families, two languages crossed with two
acquisition contexts. The English intraoral scan family spans 1{,}496 reports over 991 cases at a
median of 95 tokens, and roughly half the cases carry two or more independent reports, permitting a
direct agreement estimate: token Jaccard 0.472 and finding Jaccard 0.469 over 498 pairs. The English
photograph family spans 907 reports over 872 cases at a median of 147 tokens, being longer because
it continues past occlusion into dental health observations. Both follow a fixed
dictation order, namely transverse, vertical, sagittal, midlines, occlusal curves and crowding, so a
rule based parser recovers the structured findings at roughly 98 percent coverage on the sagittal
fields, turning the corpus into supervision.

\paragraph{The scorer as deployed:}
\looseness=-1
The published ranking weights RadFact F1 at 0.8 and the mean of BLEU-4 and METEOR at 0.2. RadFact~\cite{bannur2024maira2} parses both reports into phrases with a language
model and tests whether each is entailed by the other, penalising unsupported assertions through
precision and omissions through recall. The captioning half is less standard than
it appears, since the evaluation container sets an environment flag routing the evaluator into local
fallbacks. BLEU-4~\cite{papineni2002bleu} becomes corpus level NLTK~\cite{bird2004nltk} scoring
with uniform four gram weights and smoothing method one, pooled over the case set rather than
averaged per case, and METEOR~\cite{banerjee2005meteor} becomes a reduced variant, denoted
METEOR-lite below, with exact token matching only, greedy first match alignment, a fragmentation
penalty $0.5(c/m)^3$ over $c$ chunks and $m$ matches, and an F-mean of $F = 10PR/(R + 9P)$ that
weights recall nine times as heavily as precision. A byte faithful reimplementation of both drives
all offline decisions.

\begin{figure}[t]
\centering
\includegraphics[width=0.66\textwidth]{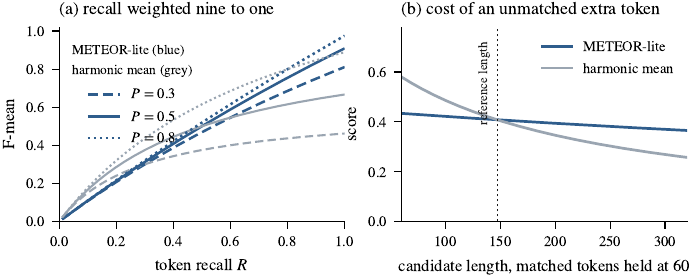}
\caption{The deployed METEOR variant. In (b), with matched tokens held at 60 against a 147 token
reference, unmatched extra tokens cost the harmonic mean 55 percent of its value but METEOR-lite
only 16 percent.}
\label{fig:scoring}
\end{figure}

\paragraph{Consequences:}
Figure~\ref{fig:scoring} makes the effect explicit: a wrong extra assertion is close to free,
whereas an omitted topic is expensive. On real intraoral scan references a constant report, taken as
the corpus medoid, reaches BLEU-4 0.4695 and METEOR 0.6245 for a captioning mean of 0.547, whereas a
genuine second clinician report scored against the first clinician's report of the same patient
reaches 0.237 and 0.527 for 0.382. The constant report thus outscores a real expert by 0.165, and
the same renderer given ground truth fields reaches 0.682, so structure dominates while correct
field values are worth a further 0.20 on top of it.

\section{Methodology}
\label{sec:method}

\begin{figure}[t]
\centering
\includegraphics[width=\textwidth]{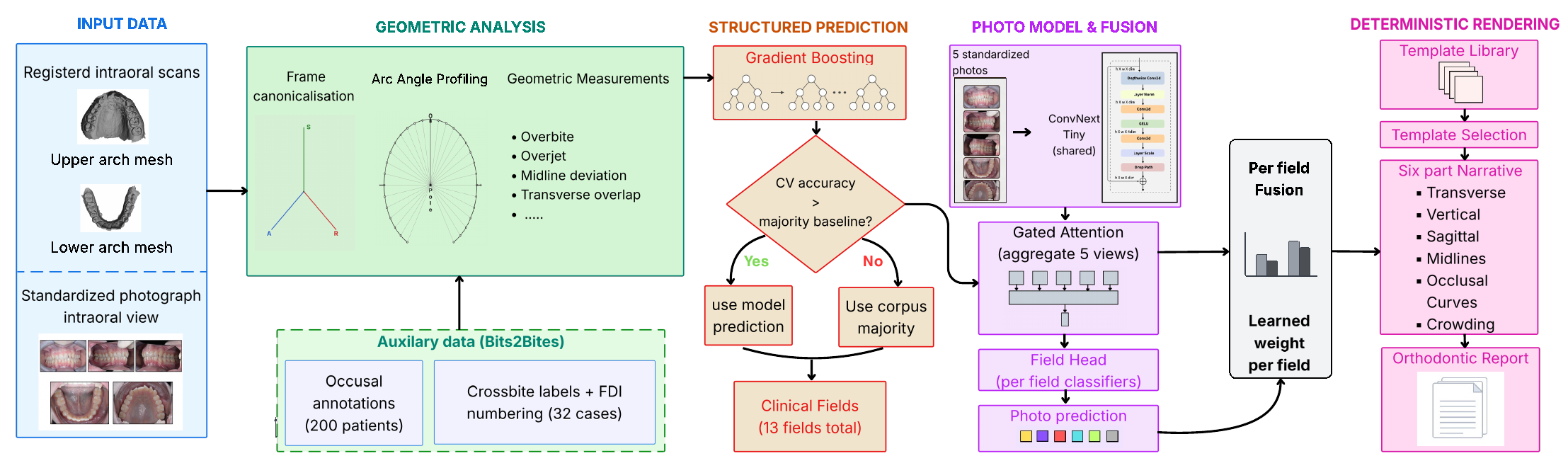}
\caption{System architecture. Geometry is always available; the photograph branch
contributes only above three distinct views. Every stage degrades to the corpus
prior rather than raising an error.}
\label{fig:architecture}
\end{figure}
\noindent

Geometry is always available, the photograph branch contributes only above three distinct views,
and every stage degrades to the corpus prior rather than raising.

\paragraph{Frame canonicalisation:}
\looseness=-1
The documentation states that scans are supplied in RAS orientation, but the release does not
conform: Bits2Bites places the anterior direction on $+y$, the Bite2Text scans place it on $-y$, and
a subset of cases places the superior axis on $y$. Assuming the stated convention corrupts every
measurement while producing no error and no log line, so the frame is recovered per case. The superior direction separates the two arch centroids; the anterior
direction follows from two cues, since a dental arch is narrower across the lateral axis at its
anterior end and closes there into a single body while splitting posteriorly into two arms. The closure cue must be evaluated on the mandible alone, because the maxillary scan includes the
palate, which fills the central region at every depth; measuring it on both arches left the median
anterior detection margin at 0.00 against approximately 2.8 after correction, costing three to ten
accuracy points on every field. Laterality is a mirror ambiguity surface geometry cannot resolve, so it is settled against the FDI
tooth numbers in the Bits2Bites crossbite annotations: over 32 informative cases the geometric
reading agrees with the RAS interpretation on 12.5 percent at a mean signed margin of $-25.6$
degrees, so $+x$ is the patient's left.

\paragraph{Arch profiling and measurement:}
\looseness=-1
A dental arch is a U shaped curve, so the natural coordinate along it is an angle. A pole inside the
arch makes every quantity a function of arch angle $\varphi$, with $\varphi=0$ at the anterior
midline and $|\varphi|$ increasing posteriorly on a one degree grid spanning $\pm150$ degrees. The
pole is the midpoint of the ridge extent rather than its median, since a U shaped point set is
bimodal laterally and the median lands on whichever arm carries more vertices, displacing the
angular origin by over a centimetre; both arches share the mandibular pole. Each angular bin yields the occlusal extreme, meaning cusp tip or incisal edge, and the buccal and
lingual limits of a 2\,mm occlusal band, and those three series give the measurements in closed
form: overbite and overjet from the incisal edges, midline deviation from the inter incisal
embrasure, the curve of Spee as chord depth along the posterior ridge, the curve of Wilson as
buccal minus lingual radial spread, transverse overlap with crossbite and scissor extent per side,
cusp interdigitation lag as the angular shift maximising cross correlation between detrended
occlusal height patterns, arch widths, and an anterior irregularity index~\cite{little1975}. Values
outside clinically possible ranges are returned as missing, so landmark failures reach the models
as absent evidence, giving 31 features per case computed with
NumPy, SciPy and trimesh~\cite{harris2020numpy,virtanen2020scipy,trimesh}.

\begin{figure}[t]
\centering
\includegraphics[width=0.68\textwidth]{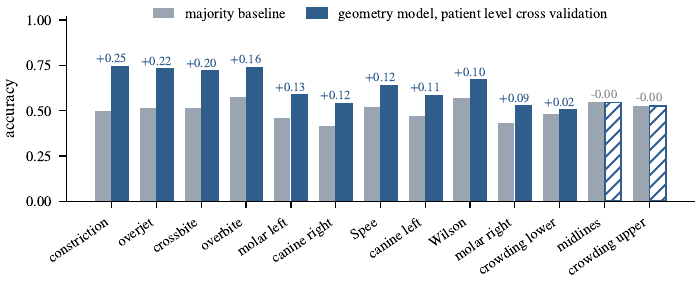}
\caption{Patient level cross validated accuracy against the per field majority baseline over 989
cases, ordered by lift. Hatched bars fail their baseline and are emitted as the corpus mode.}
\label{fig:fields}
\end{figure}

\paragraph{Field prediction:}
\looseness=-1
Supervision comes from parsing the narratives into 13 template fields, and a histogram based
gradient boosting classifier~\cite{ke2017lightgbm,pedregosa2011sklearn} is fitted per field with 400
iterations, learning rate 0.05, maximum depth three, minimum 15 samples per leaf and $L_2$
regularisation 1.0, classes under 12 supporting cases being collapsed into a bucket never emitted. The decisive choice is selective use: a field is predicted by its model only where patient
level cross validated accuracy exceeds that field's own majority baseline, and otherwise the
corpus mode is emitted~\cite{arun2026temporal}. In Figure~\ref{fig:fields}, transverse constriction, overjet, crossbite and
overbite gain 0.16 to 0.25 accuracy and the sagittal classes 0.09 to 0.13, while midlines and upper
arch crowding fail at $-0.002$ each and ship as priors. The rationale is asymmetric cost: RadFact
charges a wrong specific claim against precision, whereas the modal claim is usually correct here.

\paragraph{Photograph model and fusion:}
\looseness=-1
 Geometry is less reliable for fields requiring fine tooth-level morphology,soft tissue landmarks or photographic context, notably dental midline assessment, so a separate classifier is trained on the photographs. The five
standardised views share one ConvNeXt-Tiny backbone~\cite{liu2022convnext,wightman2019timm} of
approximately 30 million parameters, view features are pooled by gated
attention~\cite{ilse2018attention}, and one linear head per field reads the pooled embedding.
Training runs 12 epochs on one A10G GPU under AdamW~\cite{loshchilov2019adamw} at learning rate
$2\times10^{-4}$, weight decay 0.05, a one cycle schedule~\cite{smith2019superconvergence} and label
smoothing 0.05. That size is deliberate, since 600 training cases cannot support a larger model and
the result must run on CPU. The two predictors lead on different fields, geometry on overjet at 0.73 against 0.69, photographs
on right canine class at 0.63 against 0.53 and on midlines at 0.57 against 0.50, so class
probabilities are blended per field with the geometry weight selected on validation and applied
unchanged to the held out split. The branch is gated off below three distinct views, since the
challenge sample ships one file per case while training used five, and padding by repetition drops
photograph field accuracy from 0.641 to 0.508, below geometry's 0.601.

\paragraph{Narrative rendering:}
Findings are rendered into the six part narrative by a deterministic template whose phrasing options
are corpus forms selected by measured score rather than preference. Both sides of the sagittal
relationship are always stated explicitly even when they agree, although 42 percent of the corpus
uses the shorter bilateral form, because the explicit form shares more four grams with the majority
phrasing and measures 0.012 captioning higher. Where the photograph family is the target,
four dental health sentences covering restorations, sealants, caries and gingival status are
appended, photograph reports devoting approximately 3.6 sentences to observations the occlusal
section never reaches.

\section{Results}
\label{sec:results}

\looseness=-1
All held out figures use one patient disjoint split, with modal fallbacks and field models fitted on
the training half only, and scores are per case means and standard deviations, matching the public
board. Clinical factuality uses the organisers' own package, prompts and arithmetic with
Qwen2.5-7B-Instruct~\cite{yang2024qwen25} as judge, so RadFact values compare configurations rather
than predict leaderboard values.

\paragraph{Validation of the measurements:}
\looseness=-1
Bite2Text supplies no numeric occlusal ground truth, so the measurements are validated against
Bits2Bites, which carries categorical occlusal annotations over 200 patients in the same modality.
Measured overbite separates the annotated bite classes in the expected direction, at $+5.6$\,mm for
deep bite, $+2.8$\,mm for normal and $+0.2$\,mm for open bite, and crossbite extent separates
crossbite from normal cases at 56 degrees against 8. The same classifier on those measurements
reaches 0.835 accuracy on anterior bite against a 0.405 majority baseline and 0.820 on the
transverse pattern against 0.700, while median lines do not beat their baseline, anticipating the
field level pattern later seen on Bite2Text.

\begin{table}[t]
\centering\footnotesize
\caption{Held out captioning, mean and standard deviation of per case scores; field models and modal
fallbacks are fitted on the training half only. The second clinician row scores a genuine report
against another clinician's report of the same patient, and is indicative only on the photograph
family, where seven test cases carry a second report.}
\label{tab:holdout}
\begin{tabular}{llcc}
\toprule
Reference family & System & BLEU-4 & METEOR \\
\midrule
Intraoral scan & prior, no geometry & $0.4302 \pm 0.1411$ & $0.6113 \pm 0.1301$ \\
$n=296$ & geometry & $\mathbf{0.4578 \pm 0.1317}$ & $\mathbf{0.6767 \pm 0.1213}$ \\
 & oracle fields & $0.5555 \pm 0.1520$ & $0.7594 \pm 0.1271$ \\
 & second clinician, $n=145$ & $0.2234 \pm 0.0773$ & $0.5137 \pm 0.0846$ \\
\midrule
Photograph & prior, no geometry & $0.2227 \pm 0.1520$ & $0.4058 \pm 0.1506$ \\
$n=260$ & geometry, occlusal only & $0.2468 \pm 0.1579$ & $0.4416 \pm 0.1557$ \\
 & geometry and dental health & $\mathbf{0.2780}$ & $\mathbf{0.5068}$ \\
 & oracle fields, occlusal only & $0.2976 \pm 0.1957$ & $0.4878 \pm 0.1887$ \\
 & second clinician, $n=7$ & $0.2591 \pm 0.3008$ & $0.5649 \pm 0.1830$ \\
\bottomrule
\end{tabular}
\end{table}

\paragraph{Held out captioning:}
\looseness=-1
Geometry improves on the uninformed prior on both families and both metrics in
Table~\ref{tab:holdout}, by 0.028 BLEU-4 and 0.065 METEOR on the scan family and by 0.024 and 0.036
on the photograph family, and the oracle row leaves roughly 0.10 BLEU-4 of headroom inside the
renderer. The second clinician row is the most informative entry: a real expert report scores 0.2234 BLEU-4
and 0.5137 METEOR against another expert report of the same patient, inside the range occupied by
the published leaderboard. The local pipeline therefore agrees with the organisers' evaluator, and
every ranked entry operates at approximately human agreement level, so differences between entries
reflect how closely each reproduces the corpus mode rather than clinical perception. The evaluation ships one reference per case without stating its family; the first submission,
targeting the scan family, measured 0.458 and 0.677 against scan references but only 0.258 and 0.446
against photograph references, and scored 0.2639 and 0.4215 on the hidden test, fixing that family
as photographic.

\begin{table}[t]
\centering\footnotesize
\caption{RadFact under a Qwen2.5-7B-Instruct judge with the organisers' prompts and arithmetic,
intraoral scan family, $n=140$. Absolute values are not comparable with the official judge, under
which the dataset paper reports 0.372 for its strongest baseline and 0.650 for human inter reader
agreement; the ordering between configurations is the usable signal.}
\label{tab:radfact}
\begin{tabular}{lccccc}
\toprule
Configuration & Precision & Recall & RadFact F1 & Captioning & Final \\
\midrule
geometry, shipped & 0.436 & 0.485 & 0.459 & \textbf{0.580} & \textbf{0.483} \\
prior, no geometry & 0.512 & 0.432 & \textbf{0.468} & 0.527 & 0.480 \\
geometry and dental health & n/a & n/a & 0.407 & 0.532 & 0.432 \\
oracle fields, $n=24$ & 0.676 & 0.734 & 0.704 & 0.676 & 0.699 \\
\bottomrule
\end{tabular}
\end{table}

\begin{figure}[t]
\centering
\includegraphics[width=0.72\textwidth]{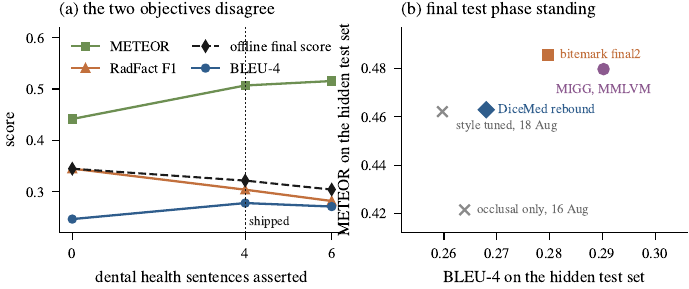}
\caption{(a) Held out effect of the dental health section on the photograph family: captioning
metrics rise with asserted content while RadFact F1 falls. (b) Final test phase standing; crosses
are two earlier submissions from the same team.}
\label{fig:tradeoff}
\end{figure}

\paragraph{Clinical factuality and objective divergence:}
\looseness=-1
In Table~\ref{tab:radfact} the uninformed prior scores marginally higher on RadFact F1 alone than
geometry, by 0.009, consistently at both sample sizes examined. The phrase counts show the
mechanism: the prior's fixed wording is segmented into 11.0 phrases per report against geometry's
7.6, and more are entailed, 5.63 against 3.46, because generic modal claims are usually true. Geometry nevertheless covers more distinct reference findings, 4.29 entailed reference phrases
against 3.85, so it is the more informative system per case while the prior wins precision through
redundancy, and it takes the combined score and the captioning board outright. The dental
health section shows the same tension in Figure~\ref{fig:tradeoff}(a): asserting no such sentence
gives RadFact F1 0.345 and captioning 0.344, four gives 0.304 and 0.392, and six gives 0.282 and
0.393. Since the public board ignores RadFact while the offline ranking weights it at 0.8, the two
objectives select different configurations, and four sentences was shipped as the configuration
attaining the highest BLEU-4 examined while forfeiting the least clinical precision.

\paragraph{Photograph fusion:}
Fusing the ConvNeXt-Tiny classifier with geometry under per field validation weights raises mean
field accuracy from 0.601 to 0.683 and lifts end to end held out scores from BLEU-4 0.2810 and
METEOR 0.5079 to 0.2930 and 0.5213. Because each weight is selected on validation and applied to a
disjoint test split, fusion cannot in expectation fall below the better single source, and the gain
arrives almost entirely on midlines and canine classification. It was completed after the test phase
closed and did not contribute to the ranked result.

\paragraph{Negative results:}
\looseness=-1
Five alternatives were measured and rejected. A vision model predicting the six dental health
findings loses to their base rates on all six, worst on gingival recession at 0.019 against 0.981,
because the labels record what a clinician chose to mention rather than what is visible and run 71
to 100 percent positive, while a model judging visibility answers negative. Retrieving the training report whose findings best match the prediction reaches 0.635 captioning
against the renderer's 0.693 even at the oracle level bounding any such retrieval, since a real
report carries variation that costs n gram overlap without adding correctness. Thresholding out of fold accuracy at
0.60, 0.65 and 0.70 to keep only the strongest fields gives captioning 0.5576, 0.5528 and 0.5504
against full geometry's 0.5801, naming FDI tooth positions lowers the worst case margin against the
leader from 1.101 to 1.051, and reordering the dental health sentences moves the objective by at
most 0.001 over 41 permutations. Finally, a phrasing variant selected by coordinate ascent measured
$+0.003$ on both metrics held out and was submitted, yet scored 0.2597 and 0.4621 on the hidden
test, a difference inside noise at 50 cases with a standard deviation near 0.15.

\section{Challenge Outcome, Deployment and Limitations}
\label{sec:outcome}

\looseness=-1
Figure~\ref{fig:tradeoff}(b) gives the final test phase standing. The submitted system placed third
at BLEU-4 $0.2680 \pm 0.1528$ and METEOR $0.4629 \pm 0.1483$, mean position 4, ranking third on
BLEU-4 and fifth on METEOR, while the two leading entries, both multimodal language model systems,
scored $0.2902 \pm 0.1756$ with $0.4797 \pm 0.1642$ and $0.2797 \pm 0.1257$ with
$0.4856 \pm 0.1407$, leaving the submitted system within 0.022 BLEU-4 of first. Held out estimates
against realised hidden test scores give calibration ratios of approximately 0.96 on BLEU-4 and 0.91
on METEOR, which project the fusion result to roughly 0.282 and 0.476.

\looseness=-1
The submitted algorithm runs CPU only, being deterministic geometry plus a gradient boosting bundle
of approximately 5\,MB. Robustness testing against a simulated platform layout covers a normal case, the flat input layout,
an empty upper mesh reproducing a real training failure, and a case with no meshes; all four exit
successfully in 3.1 to 9.3 seconds. One deployment defect deserves
recording because its cause is not obvious: the first working image needed 110 seconds per case, 104
of them inside eleven single row prediction calls, because prediction dispatches through OpenMP and
on a many core host the thread setup for a one row input dwarfs the arithmetic. Pinning to one
thread reduces a case to 6 to 8 seconds.

Three limitations bound the results. The sagittal classes remain the weakest fields at 0.53 to 0.59
accuracy, because Angle classification depends on where an individual cusp sits relative to an
individual groove, which arch level measurements only approximate, so closing that gap plausibly
requires per tooth segmentation \cite{arun2026nnunet}. RadFact figures use a substitute judge and are comparable only
among themselves, the photograph family human anchor rests on seven paired cases, and the laterality
convention comes from 32 cases in an auxiliary archive.

\section{Clinical Inference}
The system is best interpreted as an automated occlusal assessment aid rather than an autonomous orthodontic diagnosis. Geometry-derived findings such as overbite, overjet, transverse relationship and occlusal curves are directly supported by registered IOS measurements, whereas tooth-level sagittal classification, midlines and soft-tissue or dental-health findings remain less reliable or modality-dependent. For the photograph branch, future work could incorporate visual explanation
methods such as Grad-CAM and Grad-CAM++ to identify image regions supporting
tooth-level and clinical predictions \cite{george2025gradcam}. Consequently, generated reports should support clinician review and documentation rather than replace comprehensive orthodontic examination \cite{arun2025guardrails}.

\section{Conclusion}

\looseness=-1
Casting Bite2Text as measurement followed by deterministic verbalisation, rather than as multimodal
captioning, yields a small CPU resident system competitive with vision language entries. The
analysis transfers as readily as the system: the deployed metric weights recall nine to one, the
ceiling is bounded by an inter clinician agreement of 0.47, and against a single reference a
structurally complete report in the corpus idiom outscores a genuine expert report. Where the
released geometry determines a finding, measuring it beats perceiving it, and where it does not a
fused photograph classifier supplies the difference.

\end{document}